\pdfoutput=1
\documentclass{article}
\def\arxivbuild{}
\PassOptionsToPackage{numbers}{natbib}
\ifdefined\arxivbuild
  \usepackage[main, final]{neurips_2026}
\else
  \usepackage[main, final]{neurips_2026}
\fi
\usepackage[utf8]{inputenc}
\usepackage[T1]{fontenc}
\usepackage{amsmath,amssymb,amsthm}
\usepackage{booktabs}
\usepackage{tabularx}
\usepackage{graphicx}
\usepackage{hyperref}
\usepackage{url}
\DeclareUrlCommand\doiurl{\urlstyle{rm}}
\newcommand{\doi}[1]{doi:~\href{https://doi.org/#1}{\doiurl{#1}}}
\usepackage{multirow}
\usepackage[table]{xcolor}
\usepackage{tikz}
\usetikzlibrary{shapes.geometric, arrows.meta, positioning, calc, patterns, decorations.pathreplacing, fit, backgrounds}
\usepackage[capitalize]{cleveref}
\usepackage{algorithm}
\usepackage{algorithmic}
\usepackage{pifont}
\usepackage{fontawesome5}
\usepackage{float}
\usepackage[normalem]{ulem}
\usepackage{caption}
\usepackage{enumitem}
\usepackage{array}
\usepackage{microtype}
\usepackage{nicefrac}
\usepackage{placeins}

\usepackage{colortbl}
\usepackage{xcolor}
\usepackage{acronym}
\usepackage{etoolbox}
\makeatletter
\ifdefined\AC@verridelabel
  \patchcmd{\AC@verridelabel}{\ltx@label{#1}}{\ltx@label{{#1}}}{}
    {\PackageWarningNoLine{paper-body}{acronym/cleveref label patch did not apply}}
\fi
\makeatother

\definecolor{highImp}{RGB}{215, 48, 39}     % strong red
\definecolor{medHigh}{RGB}{244, 109, 67}
\definecolor{medium}{RGB}{253, 174, 97}
\definecolor{medLow}{RGB}{254, 224, 139}
\definecolor{lowImp}{RGB}{166, 217, 106}    % green
\definecolor{veryLow}{RGB}{26, 152, 80}     % strong green

\definecolor{lightgray}{gray}{0.9}

\begin{document}

% Title
\title{FTerViT: Fully Ternary Vision Transformer}

% Author
% Author
\author{
  Szymon Ruci\'nski$^{1,2}$ \quad Pietro Bonazzi$^{2}$ \\[2pt]
  \texttt{szymon.rucinski@csem.ch} \quad \texttt{pbonazzi@ethz.ch} \\[4pt]
  Engin T\"uretken$^{1}$ \quad Simon Narduzzi$^{1}$ \quad Michele Magno$^{2}$ \quad Nadim Maamari$^{1}$ \\[2pt]
  $^{1}$CSEM, Neuch\^{a}tel, Switzerland \quad $^{2}$ETH Z\"urich, Z\"urich, Switzerland
}
\maketitle
\raggedbottom

\acrodef{CE}[CE]{cross-entropy}
\acrodef{FO}[FO]{first-order}
\acrodef{KD}[KD]{knowledge-distillation}
\acrodef{LR}[LR]{learning rate}
\acrodef{MCU}[MCU]{microcontroller unit}
\acrodef{QAD}[QAD]{quantization-aware distillation}
\acrodef{QAT}[QAT]{quantization-aware training}
\acrodef{STE}[STE]{straight-through estimator}
\acrodef{ViT}[ViT]{Vision Transformer}
% =============================================================================
% ABSTRACT (What we do, why we do it?)
% =============================================================================
\begin{abstract}
Ternary Vision Transformers offer substantial model compression, however state-of-the-art methods only ternarize the encoder layers, leaving patch embeddings, LayerNorm parameters, and classifier heads in full precision. In compact models targeting resource-constrained processors, such as microcontrollers, these remaining full-precision components determine the total memory footprint, severely limiting deployment efficiency and on-device feasibility.
In this work, we introduce a fully ternarized Vision Transformer in which \emph{all} weight matrices and normalization parameters are ternarized, while activations remain 8-bit (FTerViT). To this end, we introduce two novel operators: TernaryBitConv2d with per-channel scaling for patch embedding and TernaryLayerNorm. FTerViT is trained using knowledge distillation, followed by a lightweight quantization-aware recovery phase. On a matched DeiT-S backbone, FTerViT reaches 77.47\% ImageNet-1K top-1 at 5.81\,MB, +3.27\,pp over TerViT. With the stronger DeiT-III-S backbone at 384$\times$384 resolution, our ternary W2A8 model achieves 82.43\% at 6.09\,MB (${\sim}$15$\times$ compression, $-$2.42\,pp vs.\ FP32). This backbone- and resolution-advantaged result exceeds prior ternary ViTs by up to 8\,pp. Finally, we demonstrate the first implementation of ternary vision transformers on a dual-core Xtensa LX7 microcontroller inside the ESP32-S3 system-on-chip. By deploying FTerViT-Small (based on DeiT-III-Small at 224$\times$224 resolution, 5.81\,MB), we achieve 79.64\% ImageNet-1K top-1 accuracy.

\end{abstract}
% =============================================================================
% 1. INTRODUCTION
% =============================================================================
\section{Introduction}
\label{sec:intro}

\leavevmode\acp{ViT}~\cite{dosovitskiy2020image,touvron2021training,touvron2022deit} are strong image classifiers, yet their substantial memory footprint makes them poorly suited for microcontroller-class devices. A standard compact \ac{ViT} like DeiT-Small~\cite{touvron2022deit} requires 88.3\,MB in FP32, while typical \acp{MCU} offer only a few megabytes of external RAM. This mismatch is particularly problematic for duty-cycled, event-triggered low-power vision applications where cloud offloading is undesirable or infeasible~\cite{bonazzi2023tinytracker}.

\begin{figure}[h!]
\centering
\includegraphics[width=\textwidth]{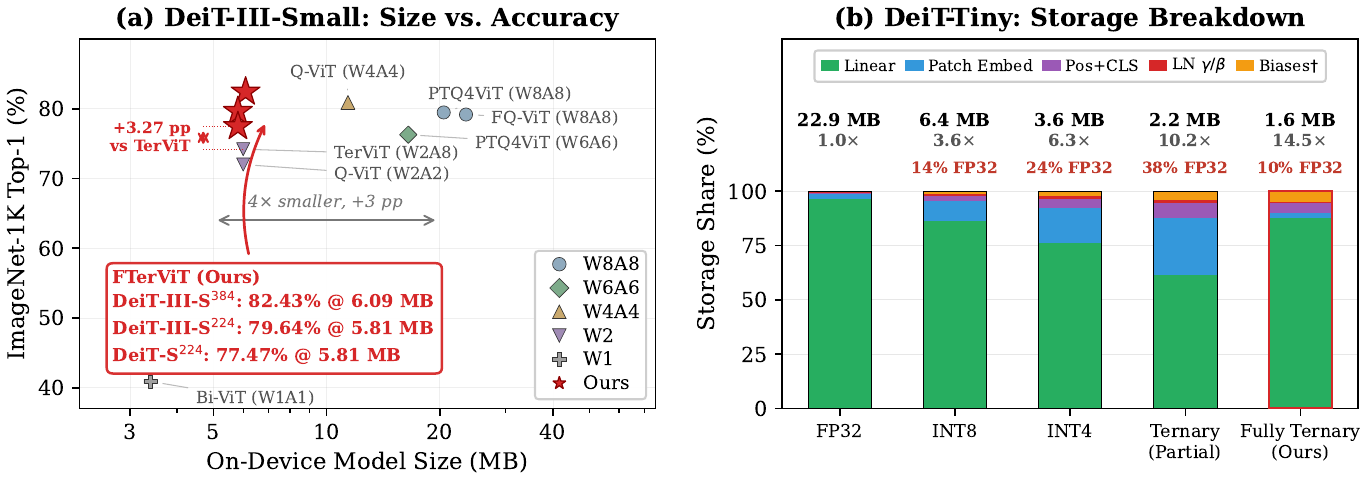}
\caption{\textbf{(a)} DeiT-III-Small size vs.\ accuracy; FTerViT (W2A8): 82.43\% at 6.09\,MB ($384^2$) / 79.64\% at 5.81\,MB ($224^2$). FTerViT based on DeiT-S$^{224}$ reaches 77.47\% under ternary quantization. \textbf{(b)} DeiT-Tiny storage; partial-W2 leaves 38\% of bytes at FP32, fully ternary drops the share to 10\%. DeiT-III-Small follows the same trend (24\% partial, 4\% fully ternary; 88.3\,MB $\to$ 5.81\,MB).}
\label{fig:teaser}
\end{figure}

Ternary quantization offers a compelling solution~\cite{yuan2024vit158b, xu2022tervit}. By constraining the weights to $\{-1,0,+1\}$ and packing them in 2 bits, ternary models can theoretically achieve 16$\times$ weight compression relative to FP32 models. However, existing low-bit \acp{ViT}~\cite{xu2022tervit,yuan2024vit158b,walczak2025bitmedvit,zhang2025ternaryclip,he2023bivit,li2024bivit,le2023binaryvit,xiao2025binaryvit,li2022qvit,liu2023ofq}, only ternarize the encoder layers, leaving the patch embedding, LayerNorm, and classifier head at INT8 or FP32. While at moderate bitwidths, these exceptions are tolerable; at 2-bits, they become the dominant bottleneck. In DeiT-Tiny~\cite{touvron2021training}, the non-ternary components account for less than 4\% of parameters yet consume 38\% of the model size (\cref{fig:teaser}b). Our proposed method ternarizes the patch embedding, LayerNorm, and classifier head jointly (\cref{tab:related_work_comparison}).

This design choice stems from well-known sensitivity of quantizing the first and last layers~\cite{huang2024quantization,li2022qvit,liu2023ofq,li2024bivit,le2023binaryvit} predating the invention of ViTs ~\cite{courbariaux2016bnn,zhou2016dorefa,dong2019hawq}. Recently, TerViT~\cite{xu2022tervit}, reports significant accuracy drops (22.4\,pp) for full-ternarization of the patch embedding and final classifier. In TerViT, LayerNorm parameters are similarly left un-quantized due to activation outliers and inter-channel variation~\cite{liu2021ptq,yuan2022ptq4vit,lin2021fqvit,li2023repqvit,yang2024dopqvit,ranjan2024lrpqvit,tai2025ampvit}. DeiT-Small~\cite{touvron2021training} showcases its patch embedding as the single most sensitive layer (4.7\,$\pm$\,0.5\% of total importance, \cref{sec:importance_results}) under a Taylor first-order analysis~\cite{molchanov2019importance}.

The resulting model substantially narrows the gap between standard ViT accuracy and MCU-scale deployment constraints. FTerViT-Small based on DeiT-III-S$^{224}$ occupies 5.81\,MB and reaches 79.64\% ImageNet-1K top-1. Based on DeiT-III-S$^{384}$, it reaches 82.43\% at 6.09\,MB, losing 2.42\,pp from FP32 and reducing the TerViT-DeiT-S accuracy gap from 5.7\,pp to 2.4\,pp (\cref{fig:teaser}a). The 224$\times$224 model runs on a \$10 ESP32-S3 with 8\,MB PSRAM, while the original 88.3\,MB FP32 model remains far beyond the device memory budget.

The main contributions of the paper are summarized as follows:

\begin{itemize}[leftmargin=*]
  \item We are the first to show that the most fragile components of \ac{ViT}: patch embedding, LayerNorms, and classifier head can be ternarized to $\{-1,0,+1\}$ when trained with knowledge distillation. Our work pushes the boundary of what was considered quantizable.
  
  \item Our approach maintains strong accuracy at extreme compression: 82.43\% top-1 on ImageNet-1K at 6.09\,MB (14.6$\times$ compression), losing 2.42\,pp from FP32. Our FTerViT-DeiT-S achieves 3.27 pp higher ImageNet-1K top-1 than TerViT-DeiT-S~\cite{xu2022tervit}.

  \item We propose a simple yet effective training strategy based on same-architecture knowledge distillation and lightweight recovery fine-tuning, enabling substantially faster convergence and improved final accuracy over prior ternary ViT training approaches.

  \item We demonstrate a standalone C implementation of a ternary Vision Transformer on the dual-core ESP32-S3, a resource-constrained commodity MCU platform with only 8MB of memory. We validate the practical feasibility of fully ternary ViTs for resource-constrained edge devices.
\end{itemize}

% =============================================================================
% 2. METHODOLOGY
% =============================================================================
\section{Methodology}
\label{sec:methodology}

FTerViT is designed to enable fully ternary Vision Transformers retaining competitive accuracy while satisfying the strict memory constraints (below 10MB) of microcontroller-class hardware. Achieving this goal is challenging because several components beyond the transformer encoder -- most notably patch embeddings and normalization layers -- are highly sensitive to quantization and are therefore typically retained in FP32 by prior work. 

This section presents the FTerViT, which is a full ternary adaptation of DeiT-Tiny~\cite{touvron2021training} (5.5M params, 22.9\,MB FP32) and DeiT-Small~\cite{touvron2022deit} (22.1M params, 88.3/88.9\,MB FP32 at 224/384 resolution) trained on ImageNet-1K. The main design of FTerViT is structured around two core elements: (1) a complete set of ternary primitives (\cref{sec:ternary_primitives}) and (2) a two-phase knowledge distillation procedure that enables stable quantization (\cref{sec:kd_recovery}). 

\subsection{Ternary Primitives: TernaryBitLinear, TernaryBitConv2d, TernaryLayerNorm}
\label{sec:ternary_primitives}

We replace the three types of weight-carrying components in a \ac{ViT} with dedicated ternary modules: TernaryBitLinear for fully-connected layers, TernaryBitConv2d for the patch-embedding convolution, and TernaryLayerNorm for LayerNorm affines. Together, these operators eliminate all remaining FP32 parameter storage while maintaining compatibility with standard \ac{ViT} architectures. The claim is fully ternary \emph{weights and normalization affines}, not ternary arithmetic: activations are quantized to 8 bits during training (\cref{eq:activation_scale}) and kept at FP32 in our MCU kernels (\cref{sec:deployment}).

\paragraph{TernaryBitLinear:}
As defined in BitNet b1.58~\cite{ma2024era}, TernaryBitLinear quantizes weights to $\{-1, 0, +1\}$ and activations to 8-bit integers. For a weight matrix $\mathbf{W} \in \mathbb{R}^{n \times m}$, the ternary quantization is:
\begin{equation}
  \tilde{\mathbf{W}} = s_w \cdot \operatorname{RoundClip}\!\left(\frac{\mathbf{W}}{s_w + \epsilon},\; -1,\; +1\right),
  \label{eq:bitlinear_quantize}
\end{equation}
where the clamped rounding operator is:
\begin{equation}
  \operatorname{RoundClip}(x, a, b) = \max\!\bigl(a,\;\min(b,\;\operatorname{round}(x))\bigr),
  \label{eq:roundclip}
\end{equation}
and $s_w$ is the per-tensor \textit{absmean} (mean absolute value) of the weight matrix:
\begin{equation}
  s_w = \operatorname{absmean}(\mathbf{W}) = \frac{1}{nm}\sum_{i,j}\bigl|\mathbf{W}_{ij}\bigr|.
  \label{eq:weight_scale}
\end{equation}
Absmean is preferred over max-based scaling because $\max|W|$ obviously pushes most ratios $|W|/s_w \ll 1$ into the zero bin of the ternarized grid. Consistent with this intuition, TWN~\cite{li2016ternary} similarly derives the optimal threshold $\Delta^* \approx 0.75\,\mathbb{E}|W|$ analytically for Gaussian weights, well below $\max|W|$. 

Following BitNet b1.58~\cite{ma2024era}, we RMS-normalize and quantize input activations of a layer to 8-bit integers via per-token absmax scaling,
\begin{equation}
  s_x = \frac{127}{\displaystyle\max_{i}\,|x_i|}\,, \qquad \tilde{x} = \operatorname{RoundClip}(s_x\,x,\,-128,\,127)\,/\,s_x.
  \label{eq:activation_scale}
\end{equation}
Gradients pass through both quantization operations via the \ac{STE}~\cite{bengio2013estimating}.

\paragraph{TernaryBitConv2d:}
For the patch embedding layer, we introduce TernaryBitConv2d, which applies per-channel scaling to better handle heterogeneous filter magnitudes~\cite{nagel2019data,krishnamoorthi2018quantizing}. For an activation map $x^{(c)} \in \mathbb{R}^{H \times W}$, let $\Omega=\{1,\ldots,H\}\times\{1,\ldots,W\}$ denote its spatial grid. Each channel $c$ has its own scales:
\begin{equation}
  s_w^{(c)} = \frac{1}{K}\sum_{k}\bigl|\mathbf{W}_{\mathrm{conv}}^{(c)}[k]\bigr|,
  \qquad
  s_x^{(c)} = \frac{127}{\displaystyle\max_{(h,w)\in\Omega}\,\bigl|x^{(c)}[h,w]\bigr|},
  \label{eq:conv_channel_scales}
\end{equation}
where $K$ is the number of kernel elements, and $H,W$ denote spatial size. The activation scale uses one scalar per sample and channel.

\paragraph{TernaryLayerNorm:}
For LayerNorm layers, we ternarize the learnable affine parameters $\gamma$ (scale) and $\beta$ (shift):
\begin{equation}
  \operatorname{TernaryLN}(\mathbf{x})
    = \tilde{\gamma} \odot \frac{\mathbf{x} - \mu}{\sqrt{\sigma^2 + \epsilon}}
    + \tilde{\beta}\,,
  \label{eq:ternary_layernorm}
\end{equation}
where $\tilde{\gamma}$ and $\tilde{\beta}$ use the same absmean scheme as weights~\eqref{eq:weight_scale}. Per-token statistics ($\mu$, $\sigma^2$), as well as biases, are kept at FP32 since their parameter count is negligible and ternarizing them would destroy positional information. The 25 LayerNorms hold $<$0.2\% of parameters but account for 34--39\% of Taylor-FO importance and 21--28\% of Hessian-trace importance (\cref{sec:importance_results}).

\subsection{Training Protocol}
\label{sec:kd_recovery}

Our ternary student is initialized directly from the pretrained FP32 teacher and trained via \ac{QAD}. Training proceeds in two phases. The first phase runs at \ac{LR} ~$1\mathrm{e}{-}4$ with cosine decay until validation top-1 saturates (reaching $76.78\%$ on DeiT-III-S$^{224}$). During the second phase, we restart \ac{LR} at~$1\mathrm{e}{-}5$ with cosine decay over 10 epochs at the same loss (see Table~\ref{tab:phase_params}). The reduction in \ac{LR} enables recalibration and recovery of the final scale from Phase~1 saturation, lifting top-1 by 3--4 \, pp across distillation settings (\cref{tab:p2_trajectory}).

During both steps, FTerViT is trained using forward-KL distillation~\cite{liu2020reactnet}, to match its pretrained frozen full-precision \ac{ViT} equivalent~\cite{he2023bivit}. Given student logits $z_S$ and teacher logits $z_T$, the FTerViT loss can be simply described as:
\begin{equation}
  \mathcal{L}_{\mathrm{KL}} = \mathrm{KL}\bigl(\mathrm{softmax}(z_T) \,\|\, \mathrm{softmax}(z_S)\bigr),
  \label{eq:ce_kl}
\end{equation}

Prior ternary and low-bit \acp{ViT} adopt a single-phase loss, ranging from label-only \ac{CE} without distillation~\cite{xu2022tervit, yuan2024vit158b}, attention or query/key similarity matching~\cite{li2022qvit, kim2022understandkdqat}, soft-logit distillation~\cite{le2023binaryvit, liu2023ofq}, hard-label distillation~\cite{xiao2025binaryvit}, multi-step \ac{KD} across bit-precisions~\cite{ranjan2024multistepkd}, to combined CE${+}$KL${+}$feature objectives~\cite{walczak2025bitmedvit}.

In contrast, we use only the KL term since a cross-entropy term conflicts with KL in low-bit networks~\cite{zhao2023sqakd,sreenivas2024minitron,xin2026qad}. In our experiments, we set the distillation temperature to $T{=}1$~\cite{wu2022tinyvit,hao2023maskedkd,sun2024logitstd}.

\begin{table}[H]
\centering
\small
\caption{Two-phase training hyperparameters. Both phases minimize only the KL loss; only the cosine learning-rate schedule changes.}
\label{tab:phase_params}
\setlength{\tabcolsep}{6pt}
\renewcommand{\arraystretch}{1.1}
\begin{tabular}{lcc}
\toprule
 & Phase 1 (Training) & Phase 2 (Fine-tuning) \\
\midrule
Loss      & $\mathcal{L}_{\mathrm{KL}}$ ($T{=}1$) & $\mathcal{L}_{\mathrm{KL}}$ ($T{=}1$) \\
Learning rate (cosine decay) & $1\mathrm{e}{-}4$ (250\,ep) & $1\mathrm{e}{-}5$ (10\,ep) \\
Optimizer & AdamW~\cite{loshchilov2019adamw}     & AdamW~\cite{loshchilov2019adamw} \\
Weight decay        & $0.05$              & $0.01$ \\
Batch     & $1024$              & $512$ \\
Teacher       & Pretrained FP32 (frozen) & Pretrained FP32 (frozen) \\
Student init  & Ternarized from teacher & Phase 1 checkpoint \\
\bottomrule
\end{tabular}
\end{table}

% =============================================================================
% 3. EXPERIMENTS
% =============================================================================
\section{Experiments}

\subsection{Layer Sensitivity Analysis}
\label{sec:importance_results}

A core obstacle to full ternarization of Vision Transformers has been the well-documented sensitivity of the patch embedding and LayerNorm layers~\cite{xu2022tervit,huang2024quantization,li2022qvit}. Prior work therefore retained these components in higher precision.

To quantify this sensitivity, we measure per-component importance on ImageNet-1K using two established estimators: Taylor first-order (FO) importance~\cite{molchanov2019importance} and Hessian-trace approximation (HAWQ-style~\cite{dong2019hawq}).

\begin{table}[H]
\centering
\small
\caption{Per-component importance share on ImageNet-1K (mean $\pm$ SEM). LayerNorm and patch embedding together dominate importance despite negligible parameter counts, explaining why prior ternary \acp{ViT} left them in higher precision. ``Top quantizable layer'' excludes non-ternarized positional/class embeddings.}
\label{tab:taylor_fo_component_share}
\setlength{\tabcolsep}{4.5pt}
\renewcommand{\arraystretch}{1.28}
\resizebox{\linewidth}{!}{%
\begin{tabular}{l cc l l c c r r}
\toprule
 & \multicolumn{3}{c}{\textbf{DeiT-Tiny (5.7\,M)}} 
 & \multicolumn{3}{c}{\textbf{DeiT-Small (22.1\,M)}} \\
\cmidrule(lr){2-4} \cmidrule(lr){5-7}
Component 
  & Params (\%) 
  & Taylor-FO (\%) 
  & Hess. (\%) 
  & Params (\%) 
  & Taylor-FO (\%) 
  & Hess. (\%) \\
\midrule
LayerNorm 
  & 0.17 & \textbf{39.4\,$\pm$\,1.2} & \textbf{28.0\,$\pm$\,0.1} 
  & 0.09 & \textbf{34.1\,$\pm$\,1.0} & \textbf{20.7\,$\pm$\,0.4} \\

FFN (FC1, FC2) 
  & 62.1 & 33.2\,$\pm$\,0.9 & 28.7\,$\pm$\,0.2 
  & 64.3 & \textbf{25.9\,$\pm$\,0.6} & 30.2\,$\pm$\,0.7 \\

Attention (Q, K, V, Proj) 
  & 31.1 & 18.4\,$\pm$\,0.4 & 15.0\,$\pm$\,0.1 
  & 32.2 & 17.4\,$\pm$\,0.3 & 19.3\,$\pm$\,0.2 \\

LayerScale 
  & \multicolumn{3}{c}{---} 
  & 0.04 & 15.7\,$\pm$\,1.2 & 9.4\,$\pm$\,0.7 \\

Patch embed 
  & 2.6 & \textbf{3.3}\,$\pm$\,\textbf{0.1} & \textbf{4.4}\,$\pm$\,\textbf{0.1}
  & 1.3 & \textbf{4.6}\,$\pm$\,\textbf{0.4} & \textbf{8.6}\,$\pm$\,\textbf{0.7} \\
Classifier head 
  & 3.4 & 1.3\,$\pm$\,0.1 & 0.2\,$\pm$\,0.0 
  & 1.7 & 0.3\,$\pm$\,0.0 & 0.01\,$\pm$\,0.00 \\
\midrule
\multicolumn{7}{c}{\faSortAmountDown~\textbf{Sorted by Taylor-FO importance (descending) on DeiT-Small}} \\
\bottomrule
\end{tabular}%
}
\end{table}

As shown in \cref{tab:taylor_fo_component_share}, all LayerNorms combined account for 21--39\% of total importance while occupying $<$0.2\% of parameters. Furthermore, among all individual layers in DeiT-Small, patch embedding is the single most important one under both metrics. These findings directly explain the 22.4\,pp accuracy drop reported by TerViT when attempting full ternarization~\cite{xu2022tervit}.

\subsection{Benchmark Results}
\label{sec:scaling_results}
We first compare FTerViT against prior quantized \acp{ViT} on ImageNet-1K. As shown in \cref{tab:related_work_comparison}, FTerViT achieves the best reported accuracy among ternary DeiT-S models, reaching \textbf{77.47\%} top-1 with 2-bit weights and 8-bit activations. A 5-seed replication of Phase 2 gives 77.45\,$\pm$\,0.07 for FTerViT-DeiT-S and 79.70\,$\pm$\,0.08 for FTerViT-DeiT-III-S$^{224}$ (mean\,$\pm$\,std), with the published single-run values inside both intervals. The seed noise is roughly 40$\times$ smaller than the +3.27\,pp margin over TerViT. Keeping the patch embedding, classifier head, and LayerNorm affines at INT8 under the identical two-phase recipe yields 78.28\% at 6.33\,MB. Full ternarization costs 0.81\,pp and buys the final 9\% of compression, so both operating points come from one unchanged recipe. Per-component, scaling-granularity, and distillation controls on the same DeiT-S backbone are reported in \cref{app:isolation_controls}.
Unlike the methods we compare against, FTerViT applies ternary quantization not only to the transformer blocks but also to the patch embedding, normalization layers, and classifier head, resulting in a fully ternary model that achieves both the best reported accuracy and the highest compression among ternary DeiT-S models.
Starting from the stronger DeiT-III-S backbone further improves performance: the $224{\times}224$ variant reaches \textbf{79.64\%} top-1 ($-3.44$\,pp from its FP32 teacher), while the $384{\times}384$ variant achieves \textbf{82.43\%} top-1 ($-2.42$\,pp).

\begin{table}[ht]
\centering
\small
\caption{Comparison of quantized \ac{ViT} methods on ImageNet-1K (Top-1). W/A = weight/activation bits. Rows with 1- or 2-bit activations (Bi-ViT, Q-ViT, LSQ) use a lower activation precision than FTerViT, their accuracies are listed for context. The ternary comparators (ViT-1.58b, TerViT) are W2A8, matching FTerViT. FTerViT is \emph{fully} ternary.}
\label{tab:related_work_comparison}

\renewcommand{\arraystretch}{1.22}
\setlength{\tabcolsep}{4pt}

\begin{tabular}{l cc l l c c r r}

\toprule
Method & W & A & Model & Size (MB) & Comp. & Regime & Epochs & Top-1 \\
\midrule 

Bi-ViT~\cite{li2024bivit} & 1 & 1 & DeiT-S & 3.4 & 26$\times$ & QAT & 300 & 40.9 \\
BiViT~\cite{he2023bivit} & 1 & mixed & Swin-S & 15.4 & 13$\times$ & QAT & 300 & 75.6 \\ 
PTQ4ViT~\cite{yuan2022ptq4vit} & 8 & 8 & DeiT-S & 22 & 4$\times$ & PTQ & 32 imgs & 79.47 \\
RepQ-ViT~\cite{li2023repqvit} & 6 & 6 & DeiT-S & 16.7 & 5.3$\times$ & PTQ & 32 imgs & 78.90\\ 
RepQ-ViT~\cite{li2023repqvit} & 4 & 4 & DeiT-S & 11 & 8$\times$ & PTQ & 32 imgs & 69.03 \\ 
Q-ViT~\cite{li2022qvit} & 2 & 2 & DeiT-S & 6.0 & 14.7$\times$ & QAT & 300 & 72.1 \\
LSQ~\cite{esser2019lsq,li2022qvit} & 2 & 2 & DeiT-S & 6.0 & 14.7$\times$ & QAT & 300 & 68.0 \\
OFQ~\cite{liu2023ofq} & 4 & 4 & DeiT-S & 11.4 & 7.7$\times$ & QAT & 325 & 81.10 \\

\midrule
\multicolumn{9}{l}{\textbf{TernaryViT}} \\
\midrule

ViT-1.58b~\cite{yuan2024vit158b} & 2 & 8 & ViT-L & 57 & 20$\times$ & Scratch & 500+ & 74.25\\ 
TerViT~\cite{xu2022tervit} & 2 & 8 & DeiT-S & 6.0 & 14.7$\times$ & QAT+PT & 300$^*$ & 74.2 \\
\rowcolor{veryLow!30}
\textbf{FTerViT (Ours)} & 2 & 8 & DeiT-S & \textbf{5.81} & \textbf{15.2$\times$} & QAD & \textbf{260} & \textbf{77.47} \\

\rowcolor{veryLow!30}
\textbf{FTerViT (Ours)} & 2 & 8 & DeiT-III-S$^{224}$ & \textbf{5.81} & \textbf{15.2$\times$} & QAD & \textbf{260} & \textbf{79.64} \\

\rowcolor{veryLow!30}
\textbf{FTerViT (Ours)} & 2 & 8 & DeiT-III-S$^{384}$ & \textbf{6.09} & \textbf{14.6$\times$} & QAD & \textbf{260} & \textbf{82.43} \\

\bottomrule
\end{tabular}

\end{table}

FTerViT uses DeiT-III-S$^{224}/^{384}$~\cite{touvron2022deit}, whose FP32 baseline (83.08--84.85\%) is higher than the DeiT-S baseline (79.86\%). Furthermore, our method generalizes well to other image classification benchmarks. As reported in Appendix~\ref{app:cifar_comparison}, our DeiT-Tiny achieves 97.43\% and 86.01\% top-1 accuracy on CIFAR-10 and CIFAR-100, respectively. These numbers are nearly on par with the full-precision DeiT-Tiny (97.52\% / 86.54\%) while using only 1.53\,MB of storage, equivalent to 15$\times$ reduction in model size.

\paragraph{Two-Phase Training Effectiveness.}
To analyze the contribution of the proposed training strategy, we study Phase~1 (high learning rate training) and Phase~2 (low learning rate recovery). As shown in Table~\ref{tab:vit_tiny_results}, while Phase~1 converges slowly toward a stable accuracy, Phase~2 achieves substantial recovery within only a few epochs.

\begin{table}[ht]
\centering
\small
\caption{FTerViT results across datasets and input resolutions. Phase~1 (250 epochs) denotes saturation performance; Phase~2 (+10 epochs) denotes final fine-tuned results.}
\label{tab:vit_tiny_results}
\setlength{\tabcolsep}{7pt}
\renewcommand{\arraystretch}{1.3}
\begin{tabular}{llc ccc}
\toprule
Model & Dataset & Input Resolution & FP32 & Phase 1 & Phase 2 \\
\midrule
DeiT-Tiny& CIFAR-10  & $224\times224$  & 97.52\% & 97.43\% & \textbf{97.43\%} \\
DeiT-Tiny & CIFAR-100 & $224\times224$  & 86.54\% & 84.42\% & \textbf{86.01\%} \\
DeiT-Tiny & ImageNet-1K & $224\times224$ & 72.19\% & 59.54\% & \textbf{63.00\%} \\
DeiT-Small & ImageNet-1K & $224\times224$ & 79.86\% & 75.05\% & \textbf{77.47\%} \\
DeiT-III-Small & ImageNet-1K & $224\times224$ & 83.08\% & 76.78\% & \textbf{79.64\%} \\
DeiT-III-Small & ImageNet-1K & $384\times384$ & 84.85\% & 78.63\% & \textbf{82.43\%} \\
\bottomrule
\end{tabular}
\end{table}

\begin{figure}[H]
    \centering
    \includegraphics[width=\linewidth]{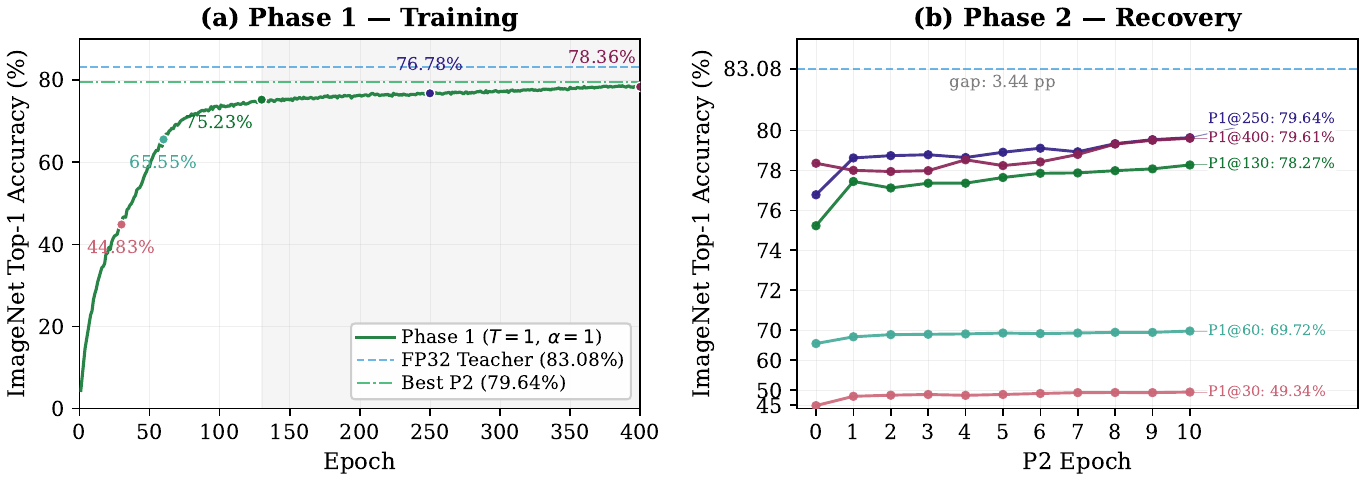}
    \caption{\textbf{(a)} Phase~1 training of DeiT-III-S$^{224}$ on ImageNet-1K for 250 epochs. Validation top-1 saturates near 78\% and never bridges the gap to FP32 (83.08\%). \textbf{(b)} Phase~2 fine-tuning from five P1 checkpoints (epochs 30--400). P1@250 converges to ${\sim}$79.64\% top-1 ($-$3.44\,pp vs.\ FP32) in 10 epochs. P1@400 reaches 79.61\%, while early checkpoints (P1@30, P1@60) recover far less.}
    \label{fig:imagenet_scaling}
\end{figure}

To better understand the source of performance gains in FTerViT, we analyze the optimization trajectory across both training phases on ImageNet-1K (DeiT-III-S$^{224}$). The key finding is that Phase~2 fine-tuning is more efficient than prolonging Phase~1: a 10-epoch low-LR restart from epoch 250 outperforms 150 additional Phase~1 epochs followed by the same restart as shown in  \cref{fig:imagenet_scaling}. More specifically, we observe the following dynamics. 

\paragraph{Phase 1: Saturation.}
Phase~1 converges to 76--78\% top-1 under cosine decay. Accuracy gains slow past epoch $130$, and extending training to epoch $400$ improves the Phase~1 checkpoint by only 1.6\,pp (76.78\% $\to$ 78.36\%).

\paragraph{Phase 2: Rapid Recovery.}
A low-\ac{LR} restart recovers accuracy within $10$ epochs regardless of Phase~1 maturity as shown in~\cref{tab:p2_trajectory}. Crucially, starting Phase~2 from epoch $250$ ($76.78\%$) reaches \textbf{79.64\%}, marginally higher than starting from epoch $400$ ($79.61\%$), demonstrating that the $150$ extra Phase~1 epochs yield no net benefit. We finetune five P1 checkpoints (epochs $30$, $60$, $130$, $250$, $400$) to confirm this pattern.

\begin{table}[ht]
\centering
\small
\caption{Phase~2 finetuning trajectory from five Phase~1 checkpoints (DeiT-III-S$^{224}$, ImageNet-1K). Same P2 recipe, \ac{LR} cosine $1\mathrm{e}{-}5\!\to\!1\mathrm{e}{-}6$ over $10$ epochs.}
\label{tab:p2_trajectory}
\setlength{\tabcolsep}{4pt}
\renewcommand{\arraystretch}{1.1}
\begin{tabular}{r r rrrrr}
\toprule
P2 Ep & LR & P1@30 & P1@60 & P1@130 & P1@250 & P1@400 \\
\midrule
0  & ---                  & 44.83 & 65.55 & 75.23 & 76.78 & \textbf{78.36} \\
1  & $9.78\mathrm{e}{-}6$ & 47.93 & 67.79 & 77.45 & \textbf{78.63} & 78.00 \\
3  & $8.15\mathrm{e}{-}6$ & 48.54 & 68.63 & 77.36 & \textbf{78.79} & 77.99 \\
5  & $5.50\mathrm{e}{-}6$ & 48.55 & 69.05 & 77.65 & \textbf{78.91} & 78.24 \\
10 & $1.00\mathrm{e}{-}6$ & 49.34 & 69.72 & 78.27 & \textbf{79.64} & 79.61 \\
\bottomrule
\end{tabular}
\end{table}

\begin{figure}[t]
    \centering

    % ===================== (a) FULL WIDTH =====================
    \begin{minipage}{\linewidth}
        \centering
        \includegraphics[width=\linewidth]{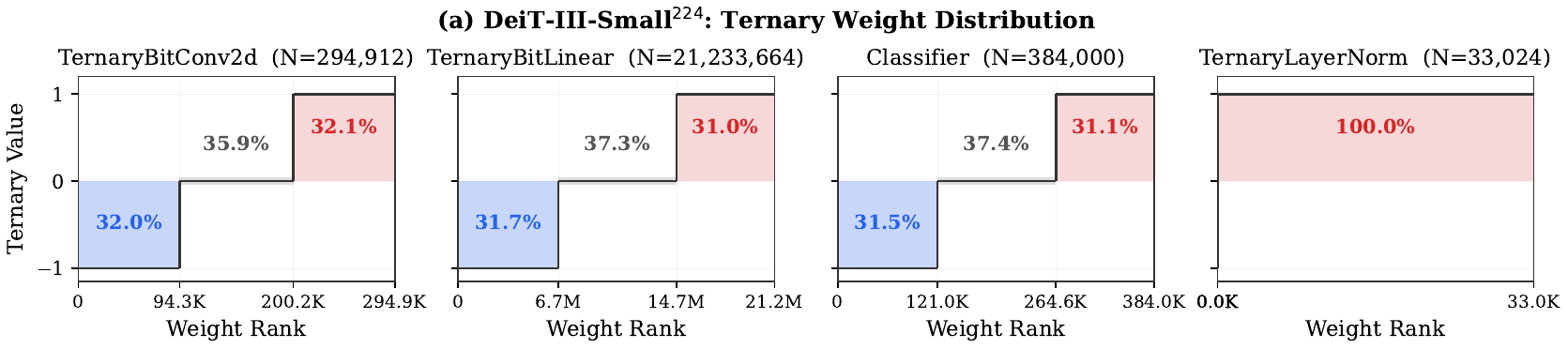}
        \vspace{2pt}

        \footnotesize
        % (a) Distribution of ternary weights across all components (DeiT-III-S$^{224}$)
    \end{minipage}

    \vspace{6pt}

    % ===================== ROW (b)(c)(d) =====================
    \begin{minipage}[t]{0.385\linewidth}
        \centering
        \includegraphics[width=\linewidth]{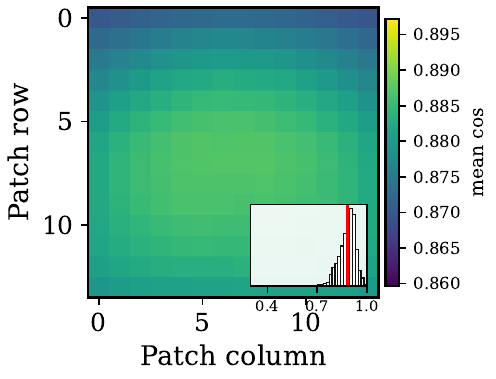}
        \vspace{2pt}

        \footnotesize
        (b) Patch embedding \\
        $\overline{\cos}=0.88 \pm 0.05$
    \end{minipage}
    \hfill
    \begin{minipage}[t]{0.294\linewidth}
        \centering
        \includegraphics[width=\linewidth]{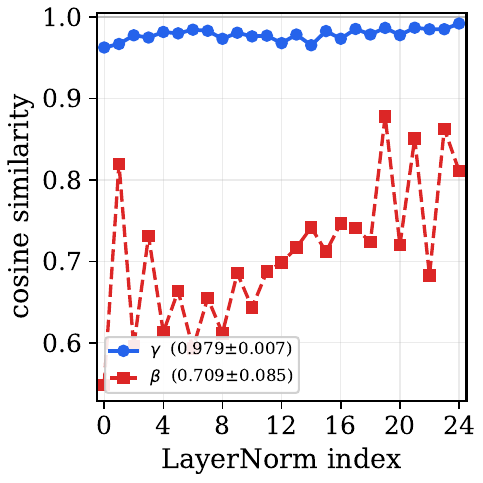}
        \vspace{2pt}

        \footnotesize
        (c) TernaryLayerNorm parameters \\
        \makebox[\linewidth][c]{$\gamma=0.979 \pm 0.007$, $\beta=0.71 \pm 0.09$}
    \end{minipage}
    \hfill
    \begin{minipage}[t]{0.302\linewidth}
        \centering
        \includegraphics[width=\linewidth]{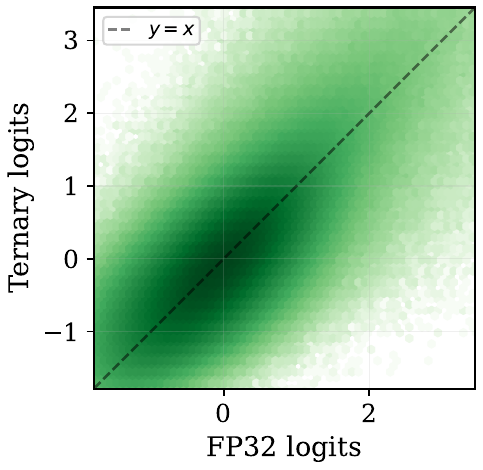}
        \vspace{2pt}

        \footnotesize
        (d) Classifier head logits \\
        $\bar{r}=0.81 \pm 0.06$
    \end{minipage}

    \caption{\textbf{Component-wise fidelity of fully ternary \acp{ViT}.}
    (a) Global distribution of weights constrained to $\{-1,0,+1\}$.
    (b,c,d) FP32–ternary alignment across key components of DeiT-III-S$^{224}$ shows strong preservation of representational structure despite ternary quantization.}

    \label{fig:component_fidelity}
\end{figure}

\subsection{Component-Level Fidelity of Ternarization}
\label{sec:component_fidelity}

We provide a detailed breakdown of ternarization effects in \cref{fig:component_fidelity}.
First, ternary weights are balanced across $\{-1,0,+1\}$, with $37.3\%$ zeros (8.18\,M weights), inducing sparsity that directly reduces multiply-accumulate operations at inference time. This distribution is consistent across TernaryBitLinear and classifier layers, while the patch embedding exhibits a slightly reduced zero fraction. At the representation level, patch embedding features remain closely aligned with FP32, with mean cosine similarity $0.88$ (std $0.05$, 5th/95th percentile $0.79$/$0.95$), indicating stable spatial feature extraction.

TernaryLayerNorm scale parameters closely match FP32 ($0.979{\pm}0.007$ cosine), while the shift is reproduced less precisely ($0.71{\pm}0.09$). The ternary $\gamma$ never uses the $-1$ level, but its per-layer absmean scale spans 0.63 to 3.86 across the 25 LayerNorms of DeiT-III-S$^{224}$, so it is not an identity transform. Setting $\gamma$ to 1 at inference collapses the trained models (77.51\% $\to$ 0.17\% for DeiT-S, 79.62\% $\to$ 4.44\% for DeiT-III-S$^{224}$). Training DeiT-S with $\gamma$ fixed to 1 from the start reaches 77.46\% vs.\ 77.47\%, so the scaling is essential but the network can learn it elsewhere when trained without $\gamma$ from the beginning.

Finally, output behavior remains consistent: classifier logits achieve mean Pearson correlation $r = 0.81$ (std $0.06$; pooled $r = 0.79$, $p < 10^{-300}$), indicating that class rankings are largely preserved despite quantization.

\subsection{CLS Token Analysis}

\begin{figure}[H]
    \centering
    \includegraphics[width=0.85\textwidth]{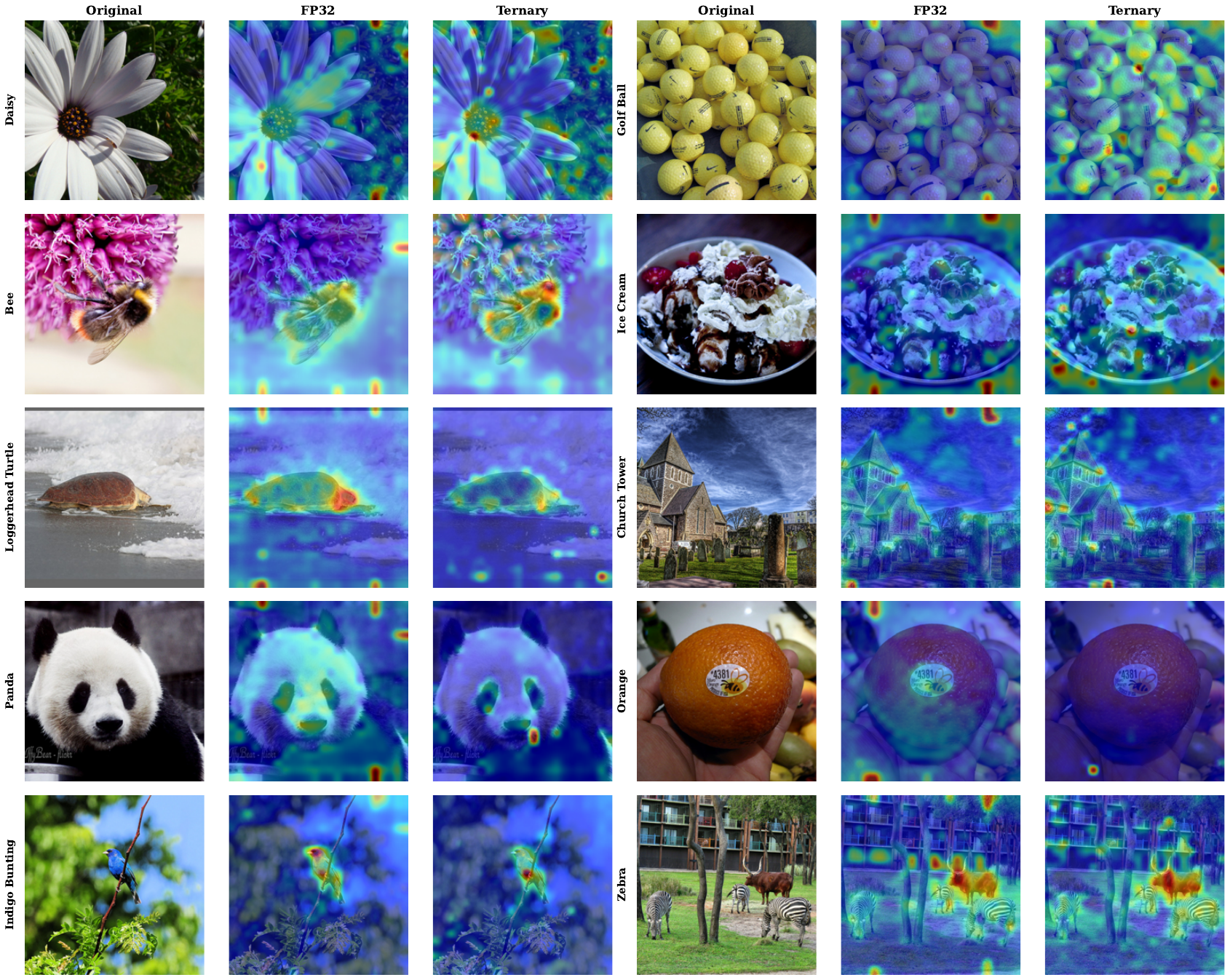}
    \caption{Attention rollout on 10 additional ImageNet-1K classes. FTerViT-DeiT-III-S$^{224}$ consistently attends to the same semantic regions as the FP32 teacher across diverse object categories.}
    \label{fig:attention_masks_extra_10}
\end{figure}

To gain deeper insights into the effectiveness of knowledge distillation for low-bit ternary models, we examine how the ternary student's internal attention patterns (FTerViT-DeiT-III-S$^{224}$) and component representations align with those of the full-precision (FP32) teacher. Attention rollout maps~\cite{abnar2020quantifying} computed from the final CLS token provide a principled way to visualize where each model directs its focus across the image. As shown in Figure~\ref{fig:attention_masks_extra_10}, the ternary student consistently attends to the same salient semantic regions as the FP32 teacher across a diverse set of ImageNet-1K classes, indicating successful transfer of high-level visual understanding despite aggressive quantization.

% =============================================================================
% 4. DEPLOYMENT
% =============================================================================
  \section{On-Device Deployment and Profiling}
  \label{sec:deployment}

\begin{figure}[ht]
  \centering
  \includegraphics[height=4.5cm]{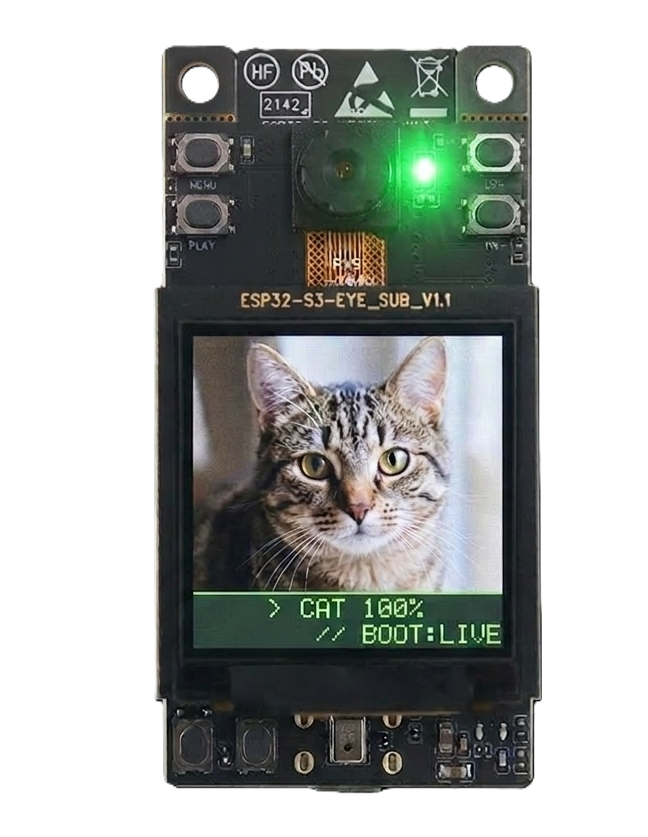}\hspace{0.3cm}
  \includegraphics[height=4.5cm]{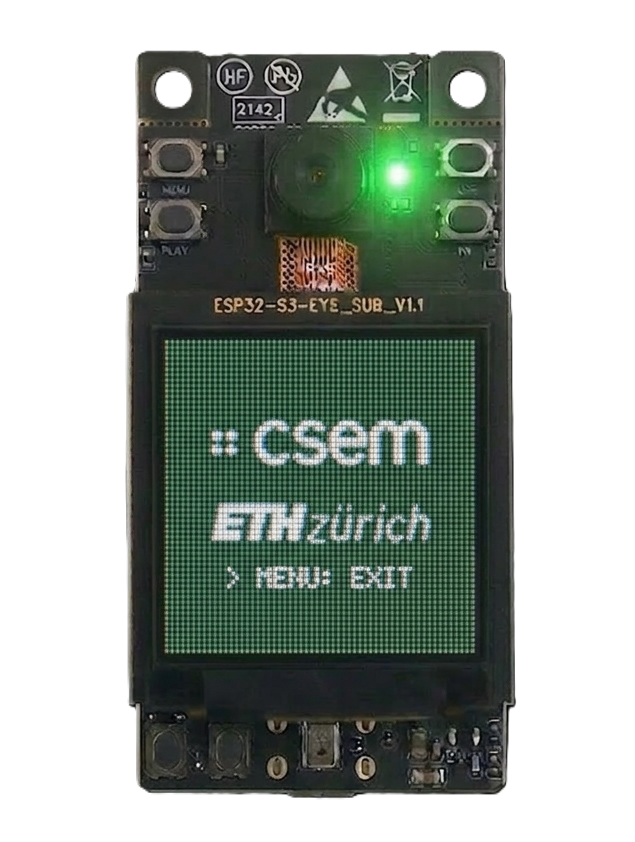}
  \caption{ESP32-S3-EYE board running on-device DeiT-III-S$^{224}$ ternary
  inference (\$10, dual-core Xtensa LX7 at 240\,MHz, 8\,MB PSRAM, 2\,MP
  camera, 240${\times}$240 LCD). The 5.81\,MB ternary model fits in flash
  (79\% partition utilization); the original 88.3\,MB FP32 checkpoint is
  $15.2\times$ larger and cannot load.}
  \label{fig:esp32_eye}
  \end{figure}

  We deploy the FTerViT based on DeiT-III-S$^{224}$ on a dual-core 32-bit LX7 microprocessor named ESP32-S3-EYE, shown in \cref{fig:esp32_eye}, running at $240$ MHz.
  Our ternary model occupies 2.83\,MB PSRAM at peak (FP32 activations and
  input image) and 5.81\,MB flash (2-bit packed weights), leaving ${\sim}4.5$\,MB PSRAM free
  on-device for camera and LCD buffers. In addition, we implement a
  standalone pure-C inference engine that executes all ternary layers without external dependencies. Ternary weights are bit-packed (4 weights per byte) into a 5.81\,MB binary, and kernels perform integer
  multiply-accumulate with fused QKV projections.

  Our ternary model can be deployed and executed entirely on-device. A
  forward pass takes $21.06$\,s, with attention (Q@K$^T$+softmax+V) and FFN
  each accounting for ${\approx}31$\% of runtime, and fused QKV projections
  contributing a further 8.3\% (see \cref{app:latency_onesp32_deit_small_224} in \cref{app:esp32_benchmark}).
  At 21\,s per inference the deployment is a feasibility result rather than a real-time system. It targets duty-cycled sensing, such as a plant-disease check from one image per node per hour or periodic state monitoring of unpowered infrastructure, where the device sleeps between inferences.

\begin{figure}[t]
\centering
\includegraphics[width=\linewidth]{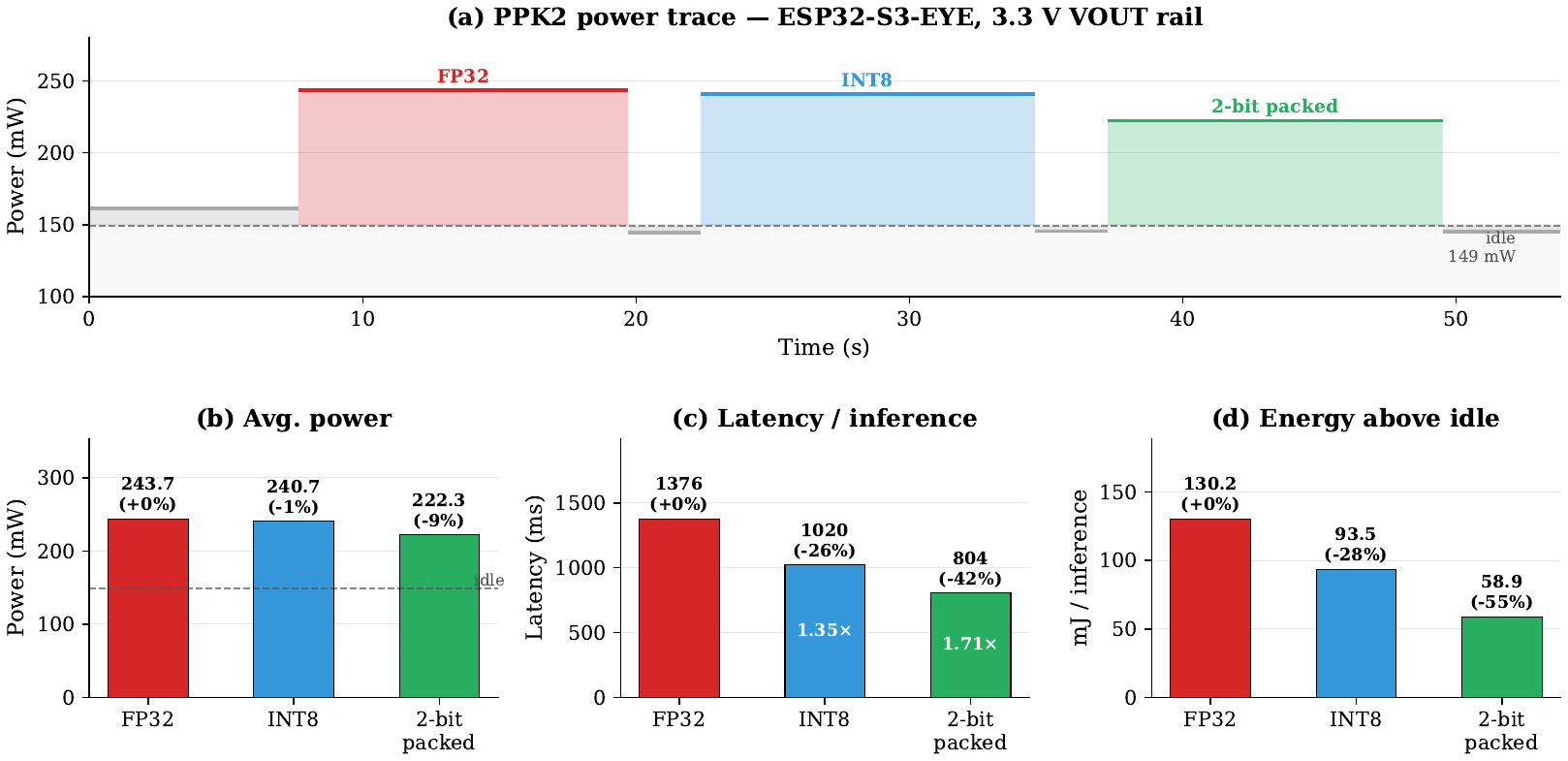}
\caption{PPK2 power measurements on ESP32-S3-EYE (Nordic PPK2 probe, VOUT rail). \textbf{Top:} raw power trace across FP32, INT8, and 2-bit packed inference phases. Dashed line = board idle (149\,mW). \textbf{Bottom:} per-format power, latency, and above-idle energy per inference. On this single FC1 kernel, packed ternary is 1.71$\times$ faster and uses 54.7\% less above-idle energy than FP32 (kernel-level, not end-to-end).}
\label{fig:ppk2_energy_main}
\end{figure}

\paragraph{Power measurements.}
We measure power on the ESP32-S3-EYE VOUT rail with a Nordic PPK2 probe (\cref{fig:ppk2_energy_main}). On the DeiT-Tiny FC1 kernel (192$\rightarrow$768), packed ternary is \textbf{1.71$\times$ faster} than FP32 (804\,ms vs.\ 1376\,ms). This is a kernel-level measurement, not end-to-end. Subtracting board idle (149\,mW), above-idle energy drops \textbf{55\%} (59\,mJ vs.\ 130\,mJ). Two effects compound: 4$\times$ smaller weight storage reduces memory-bandwidth pressure, explaining the lower latency; reduced data movement lowers active power (222\,mW vs.\ 244\,mW, ${\approx}{-}9\%$). Active-compute power falls \textbf{22.6\%} (94.8\,mW $\rightarrow$ 73.4\,mW).

\paragraph{Comparison with prior \ac{MCU} implementation of \ac{ViT}.}
Prior \ac{MCU}-scale \aclp{ViT} typically rely on neural architecture search combined with INT8 quantization to meet memory constraints. For example, MCUFormer~\cite{liang2023mcuformer} achieves 73.62\% on ImageNet-1K at 0.90\,MB, TinyFormer~\cite{yang2023tinyformer} reaches 96.10\% on CIFAR-10 at 0.91 MB, and LMaNet-Elite~\cite{elzeinaty2025tinyml} reports 74.50\% on CIFAR-100 under 1 MB. In contrast, FTerViT follows an orthogonal approach: instead of redesigning the architecture, we compress DeiT-III-S$^{224}$ from 88.3\,MB to 5.81\,MB via ternarization, achieving 79.64\% on ImageNet-1K (+6.02 pp over MCUFormer). INT8 MobileNets fit the same device at lower accuracy: MobileNetV3-Large reaches 73.8\% at 5.4\,MB and MobileNetV3-Small 64.9\% at 2.5\,MB~\cite{howard2019mobilenetv3}, while MobileNetV4-Conv-S reaches 73.8\% at FP32 (3.8\,MB INT8) and the larger V4-Conv-M/L (9.2 and 31\,MB) exceed the 8\,MB budget~\cite{qin2024mobilenetv4}.

% =============================================================================
% 5. CONCLUSION
% =============================================================================
\section{Conclusion}

FTerViT shows that all weight matrices and normalization parameters in a \ac{ViT} can be constrained to $\{-1, 0, +1\}$ with minimal accuracy loss compared to FP32. One finding stands out: \ac{KD} from a same-architecture teacher can fully ternarize \ac{ViT} architecture's most sensitive components like the patch embedding, layernorm and classifier head that prior work found extremely sensitive. Our compression pipeline's primitives (like TernaryBitLinear, TernaryBitConv2d, TernaryLayerNorm) can also be used as standalone components in novel lightweight architectures. We show that the gap to FP32 scales inversely with model capacity~\cite{xu2024lowbit}: 9.19\,pp for DeiT-Tiny vs.\ 2.42\,pp for DeiT-III-S$^{384}$. In ImageNet-1K, FTerViT achieves 82.43\% at 6.09 \, MB, surpassing previous ternary \acp{ViT} at higher compression, while 5.81\,MB DeiT-III-S$^{224}$ deploys on a \$10 ESP32-S3.

\paragraph{Limitations.} We evaluate DeiT-Tiny and DeiT-Small. Larger models exceed our \ac{MCU} memory budget, so we leave scaling untested. The C inference kernel uses basic bit-unpacking without optimization. The two-stage pipeline (training + fine-tuning) could potentially be unified into a single pass.

% =============================================================================

\begin{ack}
This research is supported by the Swiss National Science Foundation (219943) and SwissChips, a national initiative led by ETH Z\"urich, EPFL, and CSEM with funding from the State Secretariat for Education, Research and Innovation (SERI) to strengthen Switzerland's semiconductor and IC-design ecosystem.
\end{ack}

% REFERENCES
% =============================================================================
\bibliographystyle{unsrtnat}
% let long URLs break at any character so reference lines are not stretched
{\def\UrlBreaks{\do\/\do\-\do\_\do\.\do\a\do\b\do\c\do\d\do\e\do\f\do\g\do\h\do\i\do\j\do\k\do\l\do\m\do\n\do\o\do\p\do\q\do\r\do\s\do\t\do\u\do\v\do\w\do\x\do\y\do\z\do\0\do\1\do\2\do\3\do\4\do\5\do\6\do\7\do\8\do\9}%
\bibliography{references}
}

\appendix

% =============================================================================
% APPENDIX A: Experiments
% =============================================================================
\section{Experiments Appendix}

\subsection{Benchmark Results on CIFAR-10 and CIFAR-100}
\label{app:cifar_comparison}

\Cref{tab:cifar_comparison} reports DeiT-Tiny on CIFAR-10 and CIFAR-100. Our ternary model reaches 97.43\% and 86.01\% top-1. The FP32 DeiT-Tiny reaches 97.52\% and 86.54\%, so the gaps are 0.09 and 0.53\,pp. Model size drops 15$\times$, from 22.9\,MB to 1.53\,MB. FTerDeiT-Tiny is more accurate than every INT8 and ternary model in the table.

\begin{table}[H]
\centering
\small
\caption{Prior work comparison on CIFAR-10 and CIFAR-100.}
\label{tab:cifar_comparison}
\renewcommand{\arraystretch}{1.1}
\setlength{\tabcolsep}{5pt}
\begin{tabularx}{\linewidth}{l *{4}{>{\centering\arraybackslash}X}}
\toprule
 & & & \multicolumn{2}{c}{Top-1 accuracy} \\
Method & Params~(M) & Size~(MB) & CIFAR-10 & CIFAR-100 \\
\midrule
\multicolumn{5}{l}{\textbf{Full-precision}} \\
DeiT-Tiny (pretrained)~\cite{touvron2021training}
    & 5.5 & 22.9 & 97.52\% & 86.54\% \\
CCT-7/3x1~\cite{hassani2021cct}
    & 3.76 & 15 & 96.5\% & --- \\
\midrule
\multicolumn{5}{l}{\textbf{INT8 quantized}} \\
TinyFormer-300K~\cite{yang2023tinyformer}
    & 0.73 & 0.91 & 96.10\% & --- \\
LL-ViT~\cite{nag2025llvit}
    & 2.5 & 1.93 & 95.5\% & 78.8\% \\
I-ViT-T~\cite{li2023ivit}
    & 5.3 & 5.3 & 95.4\% & 79.2\% \\
MobileNetV2~\cite{sandler2018mobilenetv2}
    & 2.2 & 2.13 & 94.61\% & --- \\
LMaNet-Elite~\cite{elzeinaty2025tinyml}
    & $<$1 & $<$1 & --- & 74.50\% \\
\midrule
\multicolumn{5}{l}{\textbf{Ternary}} \\
ViT-1.58b~\cite{yuan2024vit158b}
    & 307 & 57 & 72.3\% & --- \\
\rowcolor{gray!15}
\textbf{FTerDeiT-Tiny (Ours)}
    & \textbf{5.5} & \textbf{1.53} & \textbf{97.43\%} & \textbf{86.01\%} \\
\bottomrule
\end{tabularx}
\end{table}

% =============================================================================
% APPENDIX B: ESP32 Deployment Details
% =============================================================================

\subsection{Isolation Controls on ImageNet-1K}
\label{app:isolation_controls}

Which part of the gain comes from the ternary operators, and which from distillation? We test this on FTerViT-DeiT-S$^{224}$. All numbers are top-1 on the ImageNet-1K validation set. \Cref{tab:isolation_controls} uses two protocols. A \emph{full run} is the published two-phase recipe. An \emph{adaptation} run starts from the released Phase~1 checkpoint, changes one factor, and fine-tunes for 20 epochs with the Phase~2 recipe. Adaptation runs are compared with a fully ternary model trained the same way (77.35\,$\pm$\,0.04, 5 seeds), not with the published 77.47\%.

\begin{table}[H]
\centering
\small
\caption{Isolation controls for FTerViT-DeiT-S$^{224}$ on ImageNet-1K. Mean\,$\pm$\,std over seeds where more than one seed was run. $\Delta$ is relative to the reference of the same protocol.}
\label{tab:isolation_controls}
\renewcommand{\arraystretch}{1.1}
\setlength{\tabcolsep}{5pt}
\begin{tabularx}{\linewidth}{l *{3}{>{\centering\arraybackslash}X}}
\toprule
Control & Protocol & Top-1 & $\Delta$~(pp) \\
\midrule
\multicolumn{4}{l}{\textbf{Patch-embedding scaling granularity}} \\
\rowcolor{gray!15}
\textbf{Per-channel (Ours), reference}
    & adaptation, 5 seeds & \textbf{77.35\,$\pm$\,0.04\%} & --- \\
Per-tensor
    & adaptation, 5 seeds & 77.27\,$\pm$\,0.05\% & $-$0.08 \\
\midrule
\multicolumn{4}{l}{\textbf{One component moved from ternary to INT8}} \\
Patch embedding at INT8
    & adaptation, 1 seed & 77.27\% & $-$0.08 \\
Classifier head at INT8
    & adaptation, 1 seed & 77.53\% & $+$0.18 \\
LayerNorm affines at INT8
    & adaptation, 1 seed & 76.77\% & $-$0.58 \\
All three at INT8 (6.33\,MB)
    & full run & 78.28\% & $+$0.81 \\
\rowcolor{gray!15}
\textbf{Fully ternary (Ours), 5.81\,MB}
    & full run & \textbf{77.47\%} & --- \\
\midrule
\multicolumn{4}{l}{\textbf{Distillation in Phase~2}} \\
\rowcolor{gray!15}
\textbf{KL to FP32 teacher (Ours)}
    & Phase~2, 5 seeds & \textbf{77.45\,$\pm$\,0.07\%} & --- \\
Cross-entropy only, no teacher
    & Phase~2, 3 seeds & 76.89\,$\pm$\,0.08\% & $-$0.56 \\
\bottomrule
\end{tabularx}
\end{table}

Per-channel scaling helps by 0.08\,pp over per-tensor scaling. This is within two seed standard deviations. No single component dominates. An INT8 patch embedding changes nothing beyond seed noise. An INT8 classifier head gains 0.18\,pp. INT8 LayerNorm affines lose 0.58\,pp, the largest change. Training all three at INT8 from the start, with the same full recipe, gives 78.28\% at 6.33\,MB. Full ternarization therefore costs 0.81\,pp and buys the last 9\% of compression (\cref{sec:scaling_results}). Removing the teacher from the 10-epoch Phase~2 restart costs 0.56\,pp, about eight times the seed noise.

We ran the same ablations on DeiT-Tiny and CIFAR-100 with the full two-phase recipe. The seed spread there is $\sigma{=}0.42$\,pp over 3 seeds. Distillation matters more in full training. Training without a teacher costs 3.17\,pp. Keeping the teacher but using its hard prediction instead of its soft output costs 3.16\,pp. We also isolate the cost of each component step by step. We start from an encoder-only ternary model and ternarize one more component at each step. Relative to the fully ternary model, an encoder-only ternary model is $+$0.63\,pp. Adding a ternary classifier head gives $+$0.44\,pp. Adding ternary LayerNorm affines gives $-$1.55\,pp. Adding the ternary patch embedding gives the fully ternary model. Full ternarization thus costs 0.63\,pp against the encoder-only model, less than a quarter of the distillation effect. The cost of ternary LayerNorm affines depends on the precision around them. Next to an INT8 patch embedding and head they are free ($+$0.07\,pp). Next to an FP32 patch embedding they are the largest single step.

\section{On-Device Deployment and Profiling}
\subsection{Latency and Memory Results}
\label{app:esp32_benchmark}

 \begin{table}[H]
  \centering
  \small
  \caption{On-device inference profile for DeiT-III-S$^{224}$ on ESP32-S3-EYE
  (dual Xtensa LX7 @ 240\,MHz, 8\,MB octal PSRAM, SIMD path)}
  \label{app:latency_onesp32_deit_small_224}
  \renewcommand{\arraystretch}{1.1}
  \begin{tabular*}{\linewidth}{@{\extracolsep{\fill}}lrrr@{}}
  \toprule
  Component & Time/block & 12 blocks & \% total \\
  \midrule
  LayerNorm (pre + post)        & 38\,ms   & 456\,ms     & 2.2\%  \\
  QKV projection (fused)        & 146\,ms  & 1{,}752\,ms & 8.3\%  \\
  Attention (Q@K$^T$+softmax+V) & 551\,ms  & 6{,}612\,ms & 31.4\% \\
  Output projection             & 70\,ms   & 840\,ms     & 4.0\%  \\
  FFN (fc1 + fc2)               & 549\,ms  & 6{,}588\,ms & 31.3\% \\
  \midrule
  \textbf{Total (12 blocks)}    &          & \textbf{16{,}248\,ms} & 77.1\% \\
  Patch embed + head + other    &          & 4{,}814\,ms & 22.9\% \\
  \midrule
  \textbf{End-to-end latency}   &          & \textbf{21{,}062\,ms} & 100\% \\
  Throughput                    &          & 0.0475 inf/s & --- \\
  \midrule
  \multicolumn{4}{l}{\textit{Memory \& storage}} \\
  Flash (weights, ternary)      &          & 5.81\,MB     & --- \\
  Flash (firmware, code+rodata) &          & 0.50\,MB     & --- \\
  Flash partition used / free   &          & 6.30 / 8.00\,MB & 79\% used \\
  Peak PSRAM (HWM, measured)    &          & \textbf{2.83\,MB} & --- \\
  \quad of which scratch        &          & 1.89\,MB     & --- \\
  \quad of which input image    &          & 0.57\,MB     & --- \\
  \quad of which camera/LCD     &          & $\approx$ 0.37\,MB & --- \\
  PSRAM heap available          &          & 7.35\,MB     & --- \\
  Internal SRAM min-free        &          & 93.9\,KB     & --- \\
  \midrule
  \multicolumn{4}{l}{\textit{Compression vs FP32}} \\
  FP32 \texttt{.pth} reference  &          & 88.3\,MB     & --- \\
  Ternary flash footprint       &          & 5.81\,MB     & --- \\
  ImageNet-1K top-1 accuracy    &          & 79.64\%       & --- \\
  \bottomrule
  \end{tabular*}
  \end{table}

\ifdefined\arxivbuild\else
% =============================================================================
% APPENDIX C: Reproducibility
% =============================================================================
\section{Reproducibility}
\label{app:reproducibility}

Each model trains in 1 to 2 days on a server with 8 NVIDIA L40S GPUs. Pretrained models and code are available at \url{https://huggingface.co/szymonrucinski/FTerViT} and \url{https://github.com/szymonrucinski/FTerViT}.
\fi

\ifdefined\arxivbuild\else
\clearpage
\input{checklist.tex}
\fi

\end{document}